\documentclass[lettersize,journal]{IEEEtran}
\usepackage{amsmath,amsfonts}
\usepackage{stfloats}
\usepackage[hidelinks]{hyperref}
\usepackage{multicol} % For multi-column layout
\usepackage{cite}
\usepackage{amsthm,mathtools}
\usepackage{graphicx}
\usepackage[table,dvipsnames]{xcolor}
\usepackage{graphics}
\usepackage{adjustbox}
\usepackage{graphicx}
\usepackage{threeparttable}
\usepackage{multirow} 
\usepackage{layouts}
\usepackage{enumitem}
\usepackage{amsmath,amsfonts}
\usepackage{array}
\usepackage[caption=false,font=normalsize,labelfont=sf,textfont=sf]{subfig}
\usepackage{textcomp}
\usepackage{stfloats}
\usepackage{url}
\usepackage{verbatim}
\usepackage{graphicx}
\def\BibTeX{{\rm B\kern-.05em{\sc i\kern-.025em b}\kern-.08em
    T\kern-.1667em\lower.7ex\hbox{E}\kern-.125emX}}
\usepackage{balance}

\usepackage{listings}
\usepackage{xcolor}    
\usepackage{subcaption} 

\usepackage{algorithm}
\usepackage{algpseudocode}

\usepackage{fontawesome}
\usepackage{amsfonts}% http://ctan.org/pkg/amssymb
\usepackage{pifont}% http://ctan.org/pkg/pifont

\colorlet{mc}{LimeGreen!50!White!50!}

\usepackage{xspace}

\newcolumntype{?}{!{\vrule width 1pt}}
\makeatletter
\newcommand{\thickhline}{%
    \noalign {\ifnum 0=`}\fi \hrule height 1pt
    \futurelet \reserved@a \@xhline
}
\newcolumntype{"}{@{\hskip\tabcolsep\vrule width 1pt\hskip\tabcolsep}}
\makeatother

\usepackage{tikz,xcolor,hyperref} % Orcid
\definecolor{lime}{HTML}{A6CE39}
\DeclareRobustCommand{\orcidicon}{
\begin{tikzpicture}
\draw[lime, fill=lime] (0,0) 
circle [radius=0.16] 
node[white] {{\fontfamily{qag}\selectfont \tiny ID}};    \draw[white, fill=white] (-0.0625,0.095) 
circle [radius=0.007];    \end{tikzpicture}
\hspace{-2mm}}
\foreach \x in {A, ..., Z}{
\expandafter\xdef\csname orcid\x\endcsname{\noexpand\href{https://orcid.org/\csname orcidauthor\x\endcsname}{\noexpand\orcidicon}}}

\begin{document}

\title{MaRVIn: A Cross-Layer ${\underline{\text{M}}}$ixed-Precision ${\underline{\text{R}}}$ISC-${\underline{\text{V}}}$ Framework for DNN ${\underline{\text{In}}}$ference, from ISA Extension to Hardware Acceleration}

\author{Giorgos Armeniakos\orcidA{},
        Alexis Maras\orcidB{},
        Sotirios Xydis\orcidC{},\\
        and Dimitrios Soudris\orcidD{},~\IEEEmembership{Member,~IEEE}% <-this % stops a space
        
\thanks{Received 27 March 2025; revised 24 July 2025; accepted 13 September 2025. \textit{(Corresponding Author: Giorgos Armeniakos)}}%
\thanks{G.~Armeniakos, A.~Maras, S.~Xydis and D.~Soudris are with the School of Electrical and Computer Engineering, National Technical University of Athens, Athens 15780, Greece (e-mail: \{armeniakos, amaras, sxydis, dsoudris\}@microlab.ntua.gr .}%
\thanks{This work is partially supported by EU Horizon research and innovation programme, under project CONVOLVE, grant agreement No 101070374.}
}

\markboth{IEEE Transactions on Computer-Aided Design of Integrated Circuits and Systems}{Armeniakos \MakeLowercase{\textit{et al.}}: MaRVIn: A Cross-Layer ${\underline{\text{M}}}$ixed-Precision ${\underline{\text{R}}}$ISC-${\underline{\text{V}}}$ Framework for DNN ${\underline{\text{In}}}$ference, from ISA Extension to Hardware Acceleration} % paper header

\newcommand{\red}[1]{{\color{red}#1}}
\newcommand{\blue}[1]{{\color{black}#1}}
\newcommand{\orange}[1]{{\color{orange}#1}}
\newcommand{\cmark}{ \ding{51}}%
\newcommand{\xmark}{ \ding{55}}%


\maketitle

\begin{abstract}
The evolution of quantization and mixed-precision techniques has unlocked new possibilities for enhancing the speed and energy efficiency of Neural Networks (NNs). 
Several recent studies indicate that adapting precision levels across different parameters can maintain accuracy comparable to full-precision models while significantly reducing computational demands. However, existing embedded microprocessors lack sufficient architectural support for efficiently executing mixed-precision NNs, both in terms of ISA extensions and hardware design. This limitation results in inefficiencies such as excessive data packing/unpacking and underutilized arithmetic units, leading to performance bottlenecks.
In this work, to address these challenges, we propose novel ISA extensions and the micro-architecture implementation specifically designed to optimize mixed-precision execution, enabling energy-efficient deep learning inference on RISC-V architectures. 
We introduce \textit{MaRVIn}, a cross-layer hardware-software co-design framework that enhances power efficiency and performance through a combination of hardware improvements, mixed-precision quantization, ISA-level optimizations, and cycle-accurate emulation.
At the hardware level, we enhance the ALU with configurable mixed-precision arithmetic (2-, 4-, and 8-bit) for weights and/or activations. To further improve execution efficiency, we employ multi-pumping to reduce execution latency and implement soft SIMD for efficient 2-bit operations.
We also extend ISA to support these mixed-precision operations.
At the software level, we integrate a pruning-aware fine-tuning method to optimize model compression. Additionally, we introduce a greedy-based design space exploration (DSE) approach to efficiently search for Pareto-optimal mixed-quantized models.
Finally, we incorporate voltage scaling to boost the power efficiency of our system.
Our extensive experimental evaluation over widely used
DNNs and datasets, such as CIFAR10 and ImageNet, demonstrates that our framework can achieve, on average, \blue{17.6}$\times$ speedup for less than 1\% accuracy loss and outperforms the ISA-agnostic state-of-the-art RISC-V cores, delivering up to 1.8 TOPs/W.
\end{abstract}


\section{Introduction}

As deep neural networks (DNNs) continue to evolve, their architectures are becoming more sophisticated to achieve higher accuracy in applications such as computer vision, speech processing, and other domains of artificial intelligence~\cite{jouppi2017datacenter}. 
The growing complexity of these models introduces significant computational demands, making it essential to adopt optimization techniques that balance efficiency and precision. One such widely embraced approach is quantization, which reduces the numerical precision of neural networks — effectively lowering the bit-width required for storage, computation, and data movement~\cite{Pulp-NN}.

Traditionally, fixed-precision formats have been the standard in both training and inference, ensuring stability and accuracy. 
However, recent insights reveal that not all neural network operations necessitate the same level of precision. 
Some computations can tolerate reduced precision without notable degradation in performance, enabling more efficient processing and lower power consumption. 
This realization has fueled interest in mixed-precision techniques~\cite{Mix-GEMM}, which dynamically adjust the bit-width based on the requirements of different layers or operations within a model. 
By allocating high precision only where necessary and reducing it elsewhere, mixed-precision quantization can significantly enhance both computational throughput and energy efficiency~\cite{csurarme}.

Such techniques are particularly valuable for resource-constrained environments, allowing even compact devices to support increasingly complex neural networks. 
Despite its potential, implementing mixed-precision execution efficiently on general-purpose CPU architectures remains a considerable challenge. 
Most Instruction Set Architectures (ISAs) provide only limited support, either restricting precision adjustments to coarse-grained levels~\cite{XpulpNN} or relying on inefficient numerical formats that introduce substantial overhead from operand packing and unpacking~\cite{Eyeriss}. 
Consequently, unlocking the full benefits of mixed-precision computing requires novel hardware-software co-design strategies that co-optimize computational efficiency while minimizing area and energy overheads of data type mismatch. 
Addressing these challenges is a key focus in modern computer architecture research, as it holds the potential to significantly enhance the performance of DNNs on resource-constrained RISC-V cores.

Despite ongoing efforts in hardware-software co-design~\cite{armeTC,armeDate22, SySMOL}, designing DNNs that strike the optimal balance between performance, accuracy, hardware cost, and model size is an intricate multi-objective challenge, especially for resource-constrained systems like TinyML~\cite{tinyml:ai}.
Optimizing one factor often compromises another-enhancing accuracy may demand more computational resources, while reducing hardware overhead can degrade model quality.
Overcoming this requires innovative strategies that tightly integrate algorithmic optimizations with hardware-aware design, pushing the boundaries of what is possible in edge AI.
\blue{Importantly, realizing the full potential of such hardware modifications also demands a flexible supporting software stack that can effectively map and optimize models for the underlying architecture.
}\label{R1C2a}

In this work, we introduce \textit{MaRVIn}, a cross-layer hardware-software co-design framework that incorporates an end-to-end flow supporting hardware enhancements, precision-flexible operations, and ISA-level optimizations.
\blue{Overall, (i) at the logic level, we introduce nine new mixed-precision configurations via ISA extensions, each activating distinct operational \textit{modes} that improve computational efficiency within the modified processor architecture.
(ii) At the software level, our framework incorporates a pruning-aware fine-tuning algorithm that reduces model complexity while maintaining high accuracy. This algorithm jointly explores different pruning ratios per layer and mixed-precision quantization schemes, enabling a fine-grained trade-off between accuracy, performance, and resource usage.}\label{R1C1}
Lastly, since circuit-level optimizations have been shown to exhibit a synergistic nature with algorithmic and logic enhancements when applied systematically, a) our framework leverages multi-pumping to allow the ALU to operate at twice the baseline frequency, enabling the execution of two operations per cycle, and b) applies a voltage scaling exploration to maximize potential power gains.

Using our framework, we elucidate the impact of a holistic cross-layer co-design exploration that is proven to outperform single-layer techniques~\cite{Shafique:DAC:2016:cross}, on designing and executing DNN models on RISC-V based processors.
Our evaluation of more than \blue{2000} mixed-precision DNN models across popular image classification datasets (MNIST, CIFAR-10, ImageNet), demonstrates that our framework, by extending the ISA of a RISC-V Ibex processor with the proposed approach, delivers an average energy efficiency of \blue{720} GOPs/W and up to 1.4 TOPs/W for a lower than 1\% top-1 accuracy degradation.
In other words, it achieves an average of \blue{17.6}$\times$ speedup compared to the original Ibex architecture.
Finally, we show that our framework offers a scalable performance based on a user’s accuracy constraint; thus, with lower quality criteria, the corresponding numbers can
increase up to 1.8 TOPs/W (for up to 4.5\% accuracy loss), outperforming significantly state-of-the-art approaches.

% \newline
\textbf{Our main contributions in this work can be summarized as follows:}

\begin{enumerate}[topsep=0pt,leftmargin=*]
\blue{\item We extend the RISC-V ISA with nine instructions, specifically designed to support various hardware optimizations. To validate the new ISA, we implement the proposed micro-architectural modifications within a modified RISC-V-based processor as a proof of concept.
    \item We propose \textit{MaRVIn}, an automated framework\footnote{Our framework is available open-source at \url{https://github.com/alexmr09/Mixed-precision-Neural-Networks-on-RISC-V-Cores}} for generating and evaluating mixed-precision DNNs in an end-to-end manner, from PyTorch to netlist. Our framework integrates a pruning mechanism and enables comprehensive evaluation through cycle-accurate RISC-V emulations.}
    \item We systematically analyze trade-offs between accuracy, speedup, and hardware cost, evaluating four widely used DNNs. Our results demonstrate significant energy efficiency gains, outperforming state-of-the-art approaches.
\end{enumerate}

\textit{MaRVIn} extends our preliminary work~\cite{iccad24mixed} by leveraging finer-granularity mixed-precision configurations to enhance flexibility and further improve operational efficiency.
In addition, greedy-based DSE strategies are introduced to quickly traverse the newly defined design space, and mixed-quantized models are obtained, optimized in multiple layers of abstraction.
Interestingly, compared to our previous version~\cite{iccad24mixed}, our average gains for less than 1\% accuracy degradation are 1.53$\times$ better, achieved with a similarly small increase in area complexity.

\section{background \& related work}

\begin{table}[t]
\setlength\tabcolsep{1.1pt}
\renewcommand{\arraystretch}{1.2}
\caption{Qualitative comparison of related works.}
\begin{threeparttable}
\begin{tabular}{c|cccccccccc|c}
\hline
\multicolumn{1}{l|}{} &
  \textbf{\cite{ottavi2020mixed}} &
  \textbf{\cite{XpulpNNV2}} &
  \textbf{\cite{lee2018unpu}} &
  \textbf{\cite{tcad20}} &
  \textbf{\cite{wang2024optimizing}} &
  \textbf{\cite{XpulpNN}} &
  \textbf{\cite{FILMQNNEF}} &
  \textbf{\cite{Mix-GEMM}} &
  \textbf{\cite{SySMOL}} &
  \textbf{\cite{10705106}} &
  Ours \\ \hline
% Low Prec.       & \textcolor{teal}{\cmark} & \textcolor{teal}{\cmark} & \textcolor{teal}{\cmark} & \xmark & \xmark & \textcolor{teal}{\cmark} & \textcolor{teal}{\cmark} & \textcolor{teal}{\cmark} & \textcolor{teal}{\cmark} & \textcolor{teal}{\cmark} \\
Mixed Prec.     & \textcolor{teal}{\cmark} & \textcolor{teal}{\cmark} & \textcolor{teal}{\cmark} & \xmark & \xmark & \textcolor{teal}{\cmark} & \textcolor{teal}{\cmark} & \textcolor{teal}{\cmark} & \textcolor{teal}{\cmark} & \textcolor{teal}{\cmark} & \textcolor{teal}{\cmark} \\
ASIP\tnote{1}            & \textcolor{teal}{\cmark} & \textcolor{teal}{\cmark} & \xmark & \xmark & \textcolor{teal}{\cmark} & \textcolor{teal}{\cmark} & \xmark & \textcolor{teal}{\cmark} & \textcolor{teal}{\cmark} & \textcolor{teal}{\cmark} & \textcolor{teal}{\cmark} \\
Fine-grained    & \xmark & \xmark & \xmark & \xmark & \xmark & \xmark & \textcolor{teal}{\cmark} & \xmark & \textcolor{teal}{\cmark} & \textcolor{teal}{\cmark} & \textcolor{teal}{\cmark} \\
HW-aware train. & \xmark & \xmark & \xmark & \xmark & \xmark & \xmark & \textcolor{teal}{\cmark} & \textcolor{teal}{\cmark} & \textcolor{teal}{\cmark} & \xmark & \textcolor{teal}{\cmark} \\
HW Optimiz.         & \textcolor{teal}{\cmark} & \xmark & \textcolor{teal}{\cmark} & \textcolor{teal}{\cmark} & \textcolor{teal}{\cmark} & \textcolor{teal}{\cmark} & \textcolor{teal}{\cmark} & \textcolor{teal}{\cmark} & \textcolor{teal}{\cmark} & \textcolor{teal}{\cmark} & \textcolor{teal}{\cmark} \\
Voltage Scaling         &\xmark & \xmark & \xmark & \xmark & \xmark & \xmark & \xmark &\xmark & \xmark & \xmark & \textcolor{teal}{\cmark} \\
Multipumping    & \xmark & \xmark & \xmark & \xmark & \xmark & \xmark & \xmark & \xmark & \xmark & \xmark & \textcolor{teal}{\cmark} \\
Packing         & \xmark & \xmark & \xmark & \xmark & \xmark & \xmark & \textcolor{teal}{\cmark} & \xmark & \xmark & \xmark &  \textcolor{teal}{\cmark} \\ \hline
\end{tabular}
\begin{tablenotes}\small
\item[] $^1$Application-Specific Instruction set Processor 
\vspace{-2ex}
\end{tablenotes}
\end{threeparttable}
\label{tab:related}\vspace{-2ex}
\end{table}

In recent years, research has focused on compression techniques like pruning and quantization~\cite{zervtc} to reduce inference time and memory usage of DNNs on edge platforms. Optimized software solutions, including ARM's CMSIS-NN~\cite{CMSIS-NN}, Google's GEMMLowp~\cite{jacob2022gemmlowp}, and STMicroelectronics' X-CUBE-AI~\cite{STM-XCUBE-AI}, along with deployment frameworks such as DORY~\cite{Dory} for RISC-V PULP processors and MCUNet~\cite{MCUNet,CMix-NN}, have gained popularity for running Quantized Neural Networks on commercial hardware. Yet, these tools remain largely restricted to 8-bit or higher precision arithmetic, limiting the potential gains of ultra-low-bit quantization.

Because mainstream processors struggle with sub-byte arithmetic, significant research has focused on designing specialized DNN accelerators~\cite{YodaNN,tcad20,Eyeriss,Barvinn,DNN_Intel_FPGA}. While these accelerators achieve notable speed and efficiency improvements, their practicality is often undermined by poor scalability, large silicon footprint, and high energy consumption.
Power consumption typically ranges from hundreds of milliwatts~\cite{YodaNN,tcad20,Eyeriss} to several watts~\cite{Barvinn}, making them unsuitable for resource-constrained environments.
Additionally, their coarse-grained precision and programming complexity limit flexibility, making them less adaptable to diverse workloads.

An alternative approach has been to extend the RISC-V ISA and design custom functional units optimized for sub-byte operations. PULP-NN~\cite{Pulp-NN} enhances DNN inference performance by introducing specialized instructions for packing and extracting smaller data vectors, such as 4-bit and 2-bit representations. However, the additional overhead introduced by these casting operations reduces the efficiency gains at lower bit-widths. XpulpNN~\cite{XpulpNN} expands support to 2, 4, and 8-bit SIMD operations but lacks native support for mixed-precision computations.

Recent works~\cite{ottavi2020mixed,lee2018unpu,Mix-GEMM,SySMOL} have explored mixed-precision arithmetic as a means to optimize accuracy, latency, and memory consumption in resource-limited environments~\cite{wang2024optimizing}. UNPU~\cite{lee2018unpu} implements bit-serial MAC units with a fixed activation size and weight precision varying between 1 and 16 bits, while~\cite{ottavi2020mixed} extends a RISC-V core with 2- and 4-bit MAC units, along with a custom controller for executing mixed-precision operations across 2, 4, and 8-bit data. Mix-GEMM~\cite{Mix-GEMM} and SySMOL~\cite{SySMOL} present co-design methodologies that integrate hardware accelerators with hardware-aware training techniques. However, Mix-GEMM considers only per-network quantization, potentially overlooking finer-grained precision tuning, whereas SySMOL restricts quantization to 1, 2, and 4-bit intra-layer precision, which may lead to accuracy degradation, particularly on complex datasets like ImageNet. Additionally, neither approach leverages low-level optimizations such as clock frequency scaling~\cite{Approximate_Computing_Survey} or the efficient packing of multiple low-precision operations within a single computational block~\cite{FILMQNNEF} to enhance throughput.

Table~\ref{tab:related} presents a qualitative comparison among key studies on RISC-V ISA extensions for mixed-precision DNNs, summarizing the discussion above. Unlike prior works, \textit{MaRVIn} uniquely introduces mixed-precision instructions that not only enable diverse computation modes but also integrate logic- and circuit-level optimizations, such as multi-pumping and soft SIMD, to enhance efficiency.
Furthermore, \textit{MaRVIn} applies hardware-aware fine-tuning to optimize model performance while considering quality constraints, offering an end-to-end framework that spans from ISA extensions to efficient hardware acceleration.
To the best of our knowledge, no existing method systematically combines these techniques. In Section~\ref{sec:soacomp}, we provide a detailed evaluation against state-of-the-art approaches.

\section{Configurable mixed-precision architecture based on Ibex core}
\label{sec:confarch}

\begin{figure}[t]
    \centering
    \includegraphics{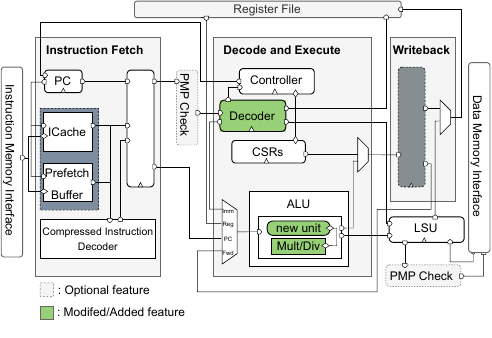}
    \caption{Ibex core architecture showcasing our modified/added features. Figure obtained from~\cite{iccad24mixed}.}
    \label{fig:ibex}
\end{figure}

The base design in this work leverages Ibex\cite{ibexproc}, a lightweight and extensible 32-bit RISC-V CPU core (Fig.\ref{fig:ibex}), widely used in energy-efficient and embedded AI applications due to its compact footprint, \blue{configurability, and active open-source development. Ibex has been silicon-proven in notable tape-outs, demonstrating its reliability and relevance in real-world deployments.
}
Without loss of generality, the specific RISC-V core serves as a reference microarchitecture scenario to showcase the impact when enabling in an execution unit the support of fine-grained mixed-precision operations, and thus can also be generalized in a straightforward manner to other RISC-V CPU cores\blue{~\cite{cv32e40p,picorv}, with potentially similar area overheads. Note that while micro-architectural modifications are Ibex-specific, our configurable mixed-precision unit and the ISA extensions are RISC-V compliant and core-agnostic.
}\label{R1C3a}

As shown in Fig.~\ref{fig:ibex}, in Ibex, the instruction fetch stage retrieves instructions from memory, which are then passed to the decode and execution stage. 
Here, the instructions are decoded to determine their operation and operands, and the ALU performs arithmetic/logic operations accordingly. 
Finally, the results are written back to the register file in the Writeback stage, completing the instruction cycle and enabling subsequent instructions to be executed.
The decode logic needs to be enhanced to support newly introduced instructions while also accommodating operations, examining both operands (i.e., weights and activations). 
This enhancement ensures proper signaling to the ALU or any newly added processing unit about the specific operation to execute.
The rest of the processor's architecture, including control logic, forwarding mechanisms, and load/store units, remains unchanged.

In the following subsections, we analyze how we modify the ALU unit to support precision-flexible operations (subsection~\ref{sec:confA}), then how these operations work and are mapped to the different respective \textit{modes} (subsection~\ref{sec:confB}), and lastly how we incorporate these \textit{modes} into new RISC-V ISA extensions (subsection~\ref{sec:confC}).

\subsection{Configurable Mixed-Precision Unit Design }\label{sec:confA}

In this section, we detail our approach to extending the RISC-V CPU core to support fine-grained mixed-precision operations, coupled with hardware optimizations, and thereby improving run-time/energy efficiency, albeit with a small, yet acceptable, degradation in accuracy.
Multiple studies have demonstrated the effectiveness of aggressive hardware storage strategies in optimizing arithmetic datapaths~\cite{SySMOL, XpulpNN}.
Hence, the key idea is to allow multiple parallel operations (i.e., MAC/MUL units) by leveraging existing logic in architectures such as multiplication units, thereby introducing minimal hardware overheads.

\begin{figure}
    \centering
    \includegraphics{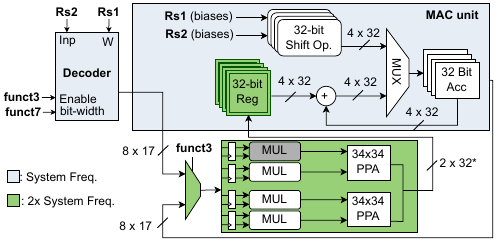}
    \caption{Microarchitecture of the modified ALU featuring packed operations, multi-pumped units, and soft SIMD optimizations. The additional multiplier is highlighted in gray. (PPA: Partial Product Addition). Figure modified from~\cite{iccad24mixed}.}
    \label{fig:micro}
\end{figure}

To achieve this, we propose and implement a modified version of Ibex's generic multiplier, expanding its capability to handle variable mixed-precision operations.
Since we primarily focus on further optimizing the performance, rather than reducing total area, we firstly select as our baseline unit the one-cycle multiplier (RV32M)~\cite{ibexproc}, featuring three parallel 17-bit $\times$ 17-bit multiplication units and a 34-bit accumulator (Fig.~\ref{fig:micro}).
Then, we expand this unit \blue{within initial ALU}\label{R1C4a} by integrating an additional 17-bit multiplier (gray MUL in Fig.~\ref{fig:micro}), allowing for parallel execution of finer-grained mixed-precision multiplications with varying bit-widths.
Interestingly, in Ibex's FPGA prototype, the additional 17-bit multiplier accommodates the existing DSPs (i.e., 4 in total) and thus introduces minimal overhead.
To perform 32-bit multiplication $A\times B$ in the Ibex core, the operands are divided into 16-bit segments with an extra bit added for carry handling, resulting in effective 17-bit values:
\begin{align}
    A &= A_{\text{high}} \| A_{\text{low}}, \\
    B &= B_{\text{high}} \| B_{\text{low}}.
\end{align}
Here, $A_{\text{high}}$ and $B_{\text{high}}$ represent the upper 16 bits, while $A_{\text{low}}$ and $B_{\text{low}}$ represent the lower 16 bits.

The multiplication is then broken into four partial products:
\begin{align}
    P_{\text{low-low}} &= A_{\text{low}} \times B_{\text{low}}, & 
    P_{\text{low-high}} &= A_{\text{low}} \times B_{\text{high}}, \\
    P_{\text{high-low}} &= A_{\text{high}} \times B_{\text{low}}, & 
    P_{\text{high-high}} &= A_{\text{high}} \times B_{\text{high}}.
\end{align}
Using these partial products, the 32-bit result is computed by combining the relevant terms:
\begin{align}
    R &= P_{\text{low-low}} + (P_{\text{low-high}} \ll 16) + (P_{\text{high-low}} \ll 16).
\end{align}
The term $P_{\text{high-high}}$ is not included in the final 32-bit result, as it represents overflow beyond the lower 32 bits. It is used only in specific instructions like MULH, where higher precision is required.
% In the subsequent clock cycle, the fourth term $P_{\text{high-high}}$ can optionally be computed if higher precision is required (e.g., for MULH).
The Ibex core uses the three parallel 17-bit multipliers to compute these partial products efficiently. 
In a single clock cycle, the three multipliers compute $P_{\text{low-low}}$, $P_{\text{low-high}}$, and $P_{\text{high-low}}$ in parallel. 
These values are then partially accumulated to form the 32-bit result.
This approach allows the core to achieve efficient 32-bit multiplication without the need for a full 32x32 multiplier. 

Furthermore, by incorporating an additional 17-bit multiplier alongside the existing three, we enhance parallelization and enable mixed-precision optimizations, such as efficient data packing, i.e., packing multiple operands together, enabling SIMD operations.
This addition preserves the original functionality of the Ibex core, requiring no modifications to the existing kernel logic, thus maintaining compatibility while improving performance and flexibility.

Our primary objective is to minimize the total required cycles of the core convolutional operation, which can be expressed as:

\begin{gather}
\textbf{Ofm}[c_o,h_o,w_o] = \sum_{c_i=0}^{C_i-1} \sum_{h_k=0}^{H_k-1} \sum_{w_k=0}^{W_k-1}
\end{gather}
\begin{gather*}
\left( \textbf{W}[c_i,c_o,h_k,w_k] \cdot \textbf{Ifm}[c_i, Sh_o+h_k, Sw_o+w_k] \right) + \textbf{b}[c_o]
\end{gather*}
\text{where:} 
\begin{align*}
&\textbf{Ifm}[C_i, H_i, W_i]: \text{input feature map} \\
&\textbf{W}[C_i, C_o, H_k, W_k]: \text{kernel weights} \\
&\textbf{b}[C_o]: \text{bias term} \\
&\textbf{Ofm}[C_o, H_o, W_o]: \text{output feature map} \\
&S: \text{stride in height (} S_h \text{) and width (} S_w \text{)}
\end{align*}

Each output feature map (Ofm) element at ($c_o,h_o,w_o$) is obtained as a weighted sum of element-wise multiplications between a local input region (Ifm) and filter weights (W), plus a bias term (b). This operation reduces to a dot product between two vectors: the extracted input values (activations $A_i$) and corresponding weights ($W_{o,i}$). 
Our approach optimizes the partial product computation $W_{o,i}\cdot A_i$, the core operation in both convolutions and fully connected layers.
Specifically, we examine 2-bit, 4-bit, and 8-bit precision for activations, alongside refining the precision of weights to 2, 4, or 8 bits.
This results in 9 distinct configurations, each of which leverages tailored optimizations to maximize efficiency while maintaining desirable quality.

\subsection{Mixed-precision Micro-architectural
Extensions}\label{sec:confB}

The baseline architecture lacks built-in precision control, requiring modifications to the existing decoder (see Fig.~\ref{fig:ibex}).
To address this, we integrate additional control logic and multiplexers, enabling dynamic operand management for low-precision data (e.g., activations and weights).
This enhanced decoder efficiently partitions operands and generates the necessary control signals, ensuring seamless execution of instructions associated with each operational \textit{mode}.

\begin{figure*}[ht]
    \centering
    \includegraphics[width=\textwidth]{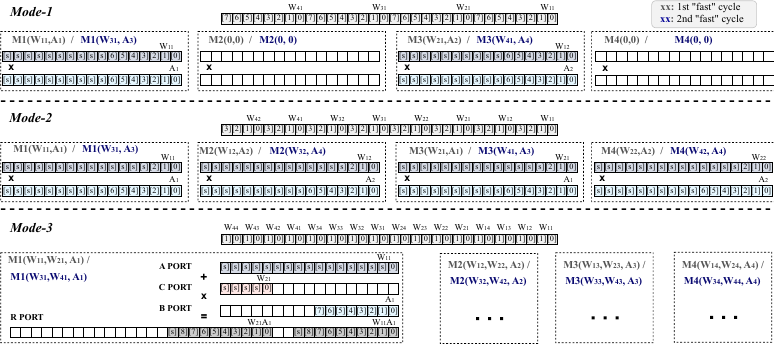}
    \caption{An illustrative example demonstrating optimized MAC operation and its mapping onto four multipliers (M1, M2, M3, M4), where each multiplier is expressed as a function of the corresponding weights ($W_{i,j}$) and activations ($A_k$). Only a partial load of weights and activations is depicted for simplicity. In \textit{Mode-1}, four 8-bit weights are packed and mapped onto two Mult units ($M_x$) along with their respective activations ($A_k$), with the remaining operations executed in the second fast cycle. In \textit{Mode-2}, eight 4-bit weights are packed and distributed across all four Mult units ($M_x$) over two fast cycles. In \textit{Mode-3}, sixteen 2-bit weights are packed, with each Mult unit ($M_k$) processing two weights per fast cycle alongside the corresponding activations.}
    \label{fig:modes}
\end{figure*}

\textbf{Mixed precision Arithmetic Logic Unit Design:}
To leverage the full potential of different low-precision configurations, we employ distinct optimization strategies at different abstraction layers.
At the logic level, given that the bit-width of weights and activations is 2, 4, or 8 bits, the initial step of the dataflow involves \textbf{data packing}, where up to 16 operands (maximally fitting 16$\times$2-bit operations) are grouped into 32-bit registers separately for weights and activations.
Each of the resulting pairs of weights and activations is mapped onto a single 17-bit $\times$ 17-bit multiplier within our modified unit.
This procedure is repeated at the same time for the remaining three multipliers.
As an example, consider performing 4 parallel partial products between eight weights, i.e.,$W_{11},W_{12},W_{21},W_{22},W_{31},W_{32},W_{41},W_{42}$, (4-bit each) and four activations, i.e., $A_1,A_2,A_3,A_4$ (8-bit each).
All weights are packed into a single 32-bit register.
During the first \textit{fast} cycle, $W_{11}A_1$ is mapped onto Mult M1, $W_{1,2}A_2$ is mapped onto Mult M2, $W_{21}A_1$ is mapped onto Mult M3 and $W_{22}A_2$ is mapped onto Mult M4.
The remaining operations are mapped accordingly in the second \textit{fast} cycle, ensuring completion of all required computations without stalls.
This example is also illustrated in Fig.~\ref{fig:modes}b, while Fig.~\ref{fig:modes}a and ~\ref{fig:modes}c depict the mapping for different operating modes.
The latter approach not only allows us to boost the parallelization of operations, but it also decreases the number of load and store instructions required by the initial (non-scaled) tactic.
Moreover, without changing the existing dataflow, the packing of activations creates a higher \textit{input reuse}~\cite{shafique:survey} opportunity.
By exploiting the sliding window mechanism, consecutive convolution operations share overlapping input data, minimizing redundant data movement (fewer memory accesses) and improving computational efficiency (reduced cycles per convolution operation).

Also at the logic level, and specifically only when 2-bit operands are used for weights, we apply a \textbf{soft SIMD} technique, making full use of the 17-bit multiplier's resources and handling as many MAC operations as possible.
\blue{
An example of this packing technique is shown in Fig.~\ref{fig:modes}c (Mode-3), where 2-bit weights and 8-bit activations are used. Each \textit{int2 $\times$ int8} multiplication yields up to a 10-bit result and thus, the second output within the accumulator must start at bit position 11 to prevent any overlap.
Additionally, for the accumulations performed in Port R during neuron computations, the required bit width must account for possible overflow
when summing multiple values.
In our case, using the formula $w = n + \log_2k$, we determine that 2 extra guard bits are needed to safely prevent overflow when accumulating $k$ $n$-bit values.
Therefore, $W_{21}$ is shifted by 12 bits (10 bits for the multiplication result plus 2 guard bits) in Port C.}\label{R2C1}
Hence, the output is calculated as follows:

\blue{
\begin{equation}
    (A_1)\cdot(W_{21}\cdot2^{12} + W_{12}) = A_1W_{21}\cdot2^{12} + A_1W_{11}\label{eq:7}
\end{equation}}\label{R2C1b}
\blue{
The first product $A_1W_{11}$ occupies the 10 least significant bits, and $A_1W_{21}$ is shifted by 12 bits to avoid overlap and ensure accuracy. This design makes efficient use of the accumulator width without additional overhead, as operand placement is automatically handled by the decoder and managed by the arithmetic units.
}

At the circuit level, motivated by the fact that smaller parallel computing units can achieve higher maximum clock frequencies, we employ \textbf{multi-pumping}.
Multi-pumping is traditionally employed to reduce resource utilization rather than execution latency~\cite{multipump:2017}. However, in this work, we exploit the fact that some functional units, such as the simpler ALU unit, have shorter critical path delays compared to more complex units. This allows us to run the ALU at twice the core clock speed, effectively executing two operations per core cycle\blue{, ensuring both clock domains are phase-aligned and derived from a common PLL, which eliminates the need for asynchronous FIFOs.}\label{R2C7}
By applying this technique to low-precision MAC operations, we accelerate the processing of packed operands: in the first \textit{fast} cycle, a subset of partial products is computed, and in the second \textit{fast} cycle, the remaining operations are completed \blue{and re-synchronized back to the slow domain through an additional register stage.} 
\blue{This ensures a continuous execution flow without additional stalls, while significantly reducing the total cycle count, primarily due to fewer weight and input memory accesses, as further discussed in the following section.}\label{R2C6a}

Summarizing, our nine mixed-precision configurations, combined with the previously discussed optimization techniques, can be grouped into three distinct modes. 
Each mode supports the same operations but differs in the bit-width of activations, offering precision levels of 2-bit, 4-bit, and 8-bit:
\begin{enumerate}[topsep=0pt,leftmargin=*]
\item $Mode\!-\!1$ (low-speed): 
if 8-bit precision is used for weights (Fig.~\ref{fig:modes}a), then multi-pumping has zero impact, since only 4 weights are loaded in one 32-bit register during the first iteration. 
Speedup is accomplished only due to multipliers parallelization and \textbf{data packing}.
% if 8-bit precision is used for weights (Fig.~\ref{fig
% }a), multi-pumping has zero impact, as only 4 weights are loaded in one 32-bit register during the first iteration. Speedup is achieved solely through multiplier parallelization.
\item $Mode\!-\!2$ (medium-speed): if 4-bit precision is used for weights (Fig.~\ref{fig:modes}b), both \textbf{multi-pumping} and data packing contribute to processing acceleration.
\item $Mode\!-\!3$ (high-speed): if 2-bit precision is used for weights (Fig.~\ref{fig:modes}c), our third \textbf{soft SIMD} optimization technique is also applied. In other words, $Mode\!-\!3$ is an optimized derivative of $Mode\!-\!2$.
\end{enumerate}

\begin{figure}
    \centering
    \includegraphics{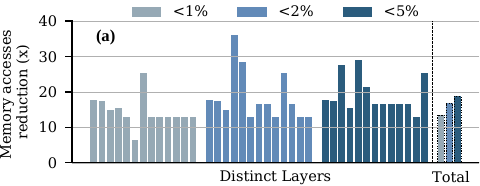}\\\vspace{1ex}
    \includegraphics{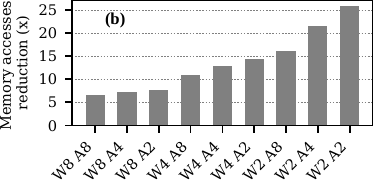}
    \caption{Memory access reduction compared to the baseline model when mapping mixed-precision operations onto the modified Ibex architecture. (a) Analysis of distinct layers from three mixed-quantized MobileNetV1 models (1\%, 2\%, 5\% accuracy loss). (b) Contribution of each configuration for a single convolutional layer.}
    \label{fig:memory}\vspace{-3ex}
\end{figure}

\subsection{RISC-V ISA Extension}\label{sec:confC}

To illustrate the speedup of our \textit{modes}, we extend the RISC-V ISA to support nine different operations, following the RISC-V ISA manual for custom extensions.
Each set of three instructions corresponds to a single mode, defined by weight and activation bit-widths, resulting in three modes and a total of nine configurations.
These nine instructions use the R-type format with six sub-fields: opcode, func3, func7, rs1, rs2, and rd, all 32 bits long. The opcode specifies the operation and the involved source (rs1, rs2) and destination (rd) registers. Func3 and func7 define the operations based on the input format.

\begin{table}[t]
\setlength\tabcolsep{1.5pt}
\renewcommand{\arraystretch}{1.2}
\footnotesize
\caption{Overview of the mixed-precision ISA extension encoding.}
\begin{tabular}{|c|c|c|c|c|c|c|}
\hline
 & \textbf{func7} & \textbf{func3} & \textbf{rs1 (A)} & \textbf{rs2 (W)} & \textbf{rd (Acc)} & \textbf{Description} \\ \hline
$nn\_mac\_w8a8$ & 000 1010 & 010 & 4 8-bit & 4 8-bit & 32-bit & Mode-1 \\ \hline
$nn\_mac\_w8a4$ & 000 0110 & 010 & 8 4-bit & 4 8-bit & 32-bit & Mode-1 \\ \hline
$nn\_mac\_w8a2$ & 000 0010 & 010 & 16 2-bit & 4 8-bit & 32-bit & Mode-1 \\ \hline
$nn\_mac\_w4a8$ & 000 1001 & 010 & 4 8-bit & 8 4-bit & 32-bit & Mode-2 \\ \hline
$nn\_mac\_w4a4$ & 000 0101 & 010 & 8 4-bit & 8 4-bit & 32-bit & Mode-2 \\ \hline
$nn\_mac\_w4a2$ & 000 0001 & 010 & 16 2-bit & 8 4-bit & 32-bit & Mode-2 \\ \hline
$nn\_mac\_w2a8$ & 000 1000 & 010 & 4 8-bit & 16 2-bit & 32-bit & Mode-3 \\ \hline
$nn\_mac\_w2a4$ & 000 0100 & 010 & 8 4-bit & 16 2-bit & 32-bit & Mode-3 \\ \hline
$nn\_mac\_w2a2$ & 000 0000 & 010 & 16 2-bit & 16 2-bit & 32-bit & Mode-3 \\ \hline
\end{tabular}
\label{tab:isa}
\end{table}

Our RISC-V mixed-precision extension instructions are listed in Table~\ref{tab:isa}. We introduce only a few instructions, having no impact on the CPU's opcode decoder hardware. 
As an example, $nn\_mac\_w8a8$ corresponds to \textit{Mode-1} and requires four 8-bit packed weights and four 8-bit activations for four parallel MAC operations.
$nn\_mac\_w4a4$ and $nn\_mac\_w2a4$ correspond to \textit{Mode-2} and \textit{Mode-3}, requiring eight 4-bit and 16 2-bit packed weights, respectively, with eight 4-bit activations. 
In all cases, the accumulator (rd) length is 32 bits. 
After completing the MAC operations for each output feature, Ibex’s standard dataflow is executed: results are retrieved from the 32-bit accumulators, followed by a requantization step to adjust the output bit-width back to 8 bits.

The key takeaways of our introduced instructions for enabling different operational \textit{Modes} are twofold.
Firstly, they facilitate a higher number of multiply operations per cycle (as shown in Section~\ref{sec:confB}), leading to increased throughput. 
Secondly, they significantly reduce the number of loads and stores, leading to fewer memory accesses, as parameters are packed and more data can be fetched in a single transfer.

For example, the reduction in memory accesses per layer (13 in total) for MobileNetV1 is depicted in Fig.\ref{fig:memory}a.
To generate Fig.\ref{fig:memory}a, we examined three mixed-precision models, ranging from high-quality models ($<$1\% accuracy loss) to less-quality ones (up to 5\%), and obtained memory accesses from Verilator~\cite{verilator}, by reporting control status of all registers.
This systematic exploration is detailed in Section~\ref{sec:framework}.
As illustrated, even with minimal quality degradation, memory accesses across different layers are reduced in total from $13.5\times$ up to almost $19\times$.
In Fig.~\ref{fig:memory}b, a deeper analysis is provided for a single convolutional layer, highlighting the impact of setting different bit-widths for weights and activations across all MAC operations within the same layer. 
Depending on the various mixed-precision configurations, the memory access reduction is shown to range from $6.5\times$ to nearly $26\times$ when aggressive precision (\textit{W2A2}) is applied.
\blue{Part of this reduction also stems from fewer pipeline stalls caused by weight/input fetches and data hazards, which account for roughly 30\% of the total execution cycles.
}\label{R2C6b}

\textbf{Compiler support:} Once the encoding of the instructions has been done, we provide a high-level interface for their integration. 
Our approach involves implementing a C intrinsic for each instruction, utilizing the inline assembly $\_\_asm\_\_$ operator to emit the bytecode corresponding to the specific instruction.
\blue{These instructions are then inserted automatically by our framework according to the model's mixed-precision configurations.}\label{R1C7}
An example with detailed instruction formats in both a baseline version and an optimized one is provided in Listings~\ref{lst:list1} and~\ref{lst:list2}, respectively.
This approach prevents us from integrating code generation within the compiler.
To facilitate the inline assembly support for our custom instructions, we made minor adjustments to the RISC-V GNU toolchain within GCC's binutils. 
This modification enables the execution of any RISC-V binary on various architectures (e.g., x86 of a host machine). 
Finally, the compiled binaries, along with the complete instruction set containing all our extensions, are executed and thoroughly emulated using the Spike simulator~\cite{spike}.

\begin{lstlisting}[language=C, caption=\small{Baseline matrix-vector multiplication in C.}, label={lst:list1}]
for (int i = 0; i < num_outputs; i++)       
    for (int j = 0; j < num_inputs; j++) 
        out[i] += w[i][j] * a[j];
\end{lstlisting}

\begin{lstlisting}[language=C, mathescape=true, caption=\small{Matrix-vector multiplication with custom ISA{.} Bold letters highlight vectorized variables{,} indicating that they contain compressed elements{.} Each variable holds four 8-bit values{,} {i.e.,} {$\textbf{out}[i] = \{ out[4i], out[4i+1], out[4i+2], out[4i+3] \}$}.}, label={lst:list2}]
for (int i = 0; i < num_outputs/4; i++) 
    for (int j = 0; j < num_inputs/4; j++) 
        asm volatile("nn_mac_w8a8 %0,%1,%2\n":
            "=r"($\textbf{out}$[i]):"r"($\textbf{a}$[j]),
            "r"($\textbf{w}$[i][4*j]):);
            ...
        asm volatile("nn_mac_w8a8 %0,%1,%2\n":
            "=r"($\textbf{out}$[i]):"r"($\textbf{a}$[j]),
            "r"($\textbf{w}$[i][4*j+3]):);
\end{lstlisting}

\section{Unified framework for cross-layer co-design exploration}\label{sec:framework}

\begin{figure}
    \centering
    \includegraphics[width=1\linewidth]{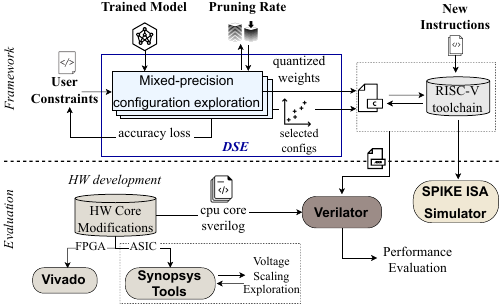}
    \caption{Overview of our proposed framework flow diagram}
    \label{fig:workflow}
\end{figure}

This section outlines the workflow of our automated framework (Fig.~\ref{fig:workflow}) for generating mixed-precision neural networks and evaluating their performance on RISC-V CPU cores.
This workflow extends our preliminary version~\cite{iccad24mixed} with three key improvements:
i) an enhanced DSE, incorporating the pruning technique, ii) support for nine new instructions, and iii) the introduction of voltage scaling exploration to boost power efficiency. 

Briefly, our framework (Fig.~\ref{fig:workflow}) receives as input a trained model (e.g., dumped from PyTorch) and performs a two-stage exploration process at the software and then at the circuit level. 
At the software level, the framework systematically explores various model configurations by first applying PyTorch's native \blue{structural} pruning method to examine different percentages of connections to prune per layer.\label{R2C3}
Following this, it evaluates fine-grained mixed-precision configurations for each layer, considering 2-bit, 4-bit, and 8-bit precision for both weights and activations. 
Once the final Pareto-optimal models are identified, logic-level optimizations are implemented by mapping operations to various operational modes based on their precision configurations.
Subsequently, the framework transitions to circuit-level exploration, applying voltage scaling to evaluate trade-offs among operating voltage, power consumption, and system frequency.
This stage also explores the available slack time and investigates how the multi-pumping technique can be fully exploited, enabling maximum potential power gains at the same time.

Once the functionality of the \textbf{nine} new instructions is simulated and verified, the linker generates a single executable file (ELF).
This file, along with the RTL descriptions of the modified core—written in SystemVerilog and synthesized using industry-standard EDA tools (Xilinx Vivado for FPGA flow and Synopsys Design Compiler for ASIC flow)—is provided as input to Verilator~\cite{verilator}, a cycle-accurate emulator.
The flow reaches the synthesis stage, but post-layout PnR validation is not performed in this flow.
Additionally, we apply voltage scaling using Synopsys tools under different operating conditions.
This setup allows for a comprehensive evaluation of performance metrics, including accuracy-speedup trade-offs, power consumption, and any introduced overheads.

\subsection{Mixed-precision exploration}\label{sec:explor}

One of our primary objectives in our work, as well as a key challenge in multi-objective optimization co-design approaches, is to identify, quite fast, a close-to-optimal design solution.
To tackle this challenge, we adopt a greedy search method (Algorithm~\ref{algo:algo}).
Note that while greedy algorithms have been proven to extract good enough solutions in complex optimization problems~\cite{macish}, any other algorithms (such as heuristics~\cite{tcad20}) can be seamlessly employed.

Our mixed-precision DSE starts with a pre-trained model $m$, where we systematically adjust different pruning rates and precision settings, which allow us to identify configurations that balance computational efficiency along with a user's user-defined accuracy loss threshold $T$.
The process begins by iterating over a range of pruning rates, defined in steps of a granularity parameter $g$, up to a maximum pruning rate $p_{max}$.
The pruning rate is simply defined as the percentage of weights per layer that are set to zero during the pruning-aware post-training process.
For each pruning rate, we analyze the impact of mixed-precision configurations on the latency of each layer in the model (line~\ref{line:calc}). 
Specifically, we evaluate all possible configurations per layer (e.g., 2-bit, 4-bit, and 8-bit precision for weights and activations) and select an arbitrary medium-latency configuration $C_l$ (line~\ref{line:conf}) for each layer as a starting point set $C_{start}$ (line~\ref{line:cstart}).
The latency of each layer can be quickly estimated using Verilator, without executing the entire model. This enables us to narrow down the design space and avoid exhaustive evaluation of all possible configurations.
The initial setting, consisting of medium-latency configurations for all layers, is then evaluated for accuracy (line~\ref{line:eval}).
If the accuracy of this starting point remains within the acceptable threshold (i.e., the accuracy loss does not exceed $T$), a full DSE is triggered (line~\ref{line:dse}). 
During this exploration, all remaining combinations of precision configurations are evaluated, excluding those with higher precision settings than the current configuration.
By pruning these higher precision configurations, we can reduce the search space and focus on solutions with lower computational costs.
If the accuracy constraint is not satisfied with the initial configuration, the algorithm identifies the layer with the lowest computational intensity in terms of required cycles and incrementally adjusts its precision to a higher setting (line~\ref{line:arg}). 
This iterative process ensures that only the necessary adjustments are made to meet the accuracy threshold while minimizing computational overhead. 
After each adjustment, the model’s accuracy is re-evaluated, and the process is repeated until the accuracy constraint is satisfied or all layers have been adjusted to their maximum allowable precision.
The entire process is terminated when DSE reaches $I$ iterations (line~\ref{line:term1} or line~\ref{line:term2}), an arbitrary number set by the user, to control iterations and prune the design space, ensuring no strict omissions of combinations between layers and configurations.
Subsequently, a set $\mathcal{D}$ of acceptable mixed-precision configurations is returned (line~\ref{line:end}).

\blue{Moreover, based on our experiments, the proposed greedy-based exploration consistently identifies solutions close to optimal. For example, in the case of LeNet, our hypervolume analysis~\cite{hypervolume} shows that it covers 73\% of the optimal trade-off space (as defined by exhaustive DSE), while achieving this with 14$\times$ less runtime. This highlights both the scalability and the practical relevance of our method, particularly for larger models where exhaustive search is infeasible.
}\label{R1C2b}

\begin{algorithm}[t!]
\caption{Pseudocode for Mixed-Precision DSE}\label{algo:algo}
\footnotesize
\textbf{\color{blue!70!black}Input:} 1) Pretrained Model $m$, 2) Accuracy Loss Threshold $T$, \\3) Granularity $g$, 4) Max Pruning Rate $p_{\text{max}}$, 5) Iterations $I$ for DSE\\
\textbf{\color{blue!70!black}Output:} Design space $\mathcal{D}^*$ with acceptable mixed-precision configurations\\
\begin{algorithmic}[1]
\State \textbf{\color{black}for} $p \in \{0, g, 2g, \dots, p_{\text{max}}\}$ {\color{gray!80!black}\Comment{\textit{Iterate over pruning rates}}}
\State \hspace{3mm} \textbf{\color{black}for} each layer $l \in m$
\State \hspace{6mm} Calculate latency for all (9) configurations of $l$\label{line:calc}
\State \hspace{3mm} \textbf{\color{black}end for}
\State \textbf{\color{black}end for}
\State \textbf{\color{black}for} each layer $l \in m$
\State \hspace{3mm} $\mathcal{C}_l \leftarrow $ medium latency config\label{line:conf}
\State \textbf{\color{black}end for}
\State $\mathcal{C}_{\text{start}} \leftarrow \{\mathcal{C}_l \mid l \in m\}$ {\color{gray!80!black}\Comment{\textit{A set of $C_l$ instances as starting point}}}\label{line:cstart}
\State Evaluate accuracy $\mathcal{A}_{\text{start}}$ of $\mathcal{C}_{\text{start}}$\label{line:eval}
\State \textbf{\color{black}if} $\mathcal{A}_{\text{start}} \geq \mathcal{A}_m - T$ {\color{gray!80!black}\Comment{\textit{Check accuracy constraint}}}
\State \hspace{3mm} \textbf{Perform full design space exploration:}\label{line:dse}
\Statex \hspace{9mm} $\mathcal{D} = \bigcup_{l \in m} \{\mathcal{C}_l \mid \text{remaining combinations}\}$
\Statex \hspace{9mm} \textit{Discard higher precision configs:}
\Statex \hspace{9mm} $\mathcal{D} \leftarrow \mathcal{D} \setminus \{\mathcal{C}_l \mid b_w, b_a \text{ are higher than current settings}\}$
\Statex \hspace{9mm} \textbf{\color{black}until} DSE reaches $I$ iterations, {\color{red!60!black}\textbf{terminate process}}\label{line:term1}
\State \textbf{\color{black}else}
\State \hspace{3mm} $l_{\text{min}} \leftarrow \arg\min_{l \in m} \text{compute\_intensity}(l)$
\State \hspace{3mm} Move to a higher configuration for $l_{\text{min}}$:\label{line:arg}
\State \hspace{6mm} $\mathcal{C}_l \leftarrow \mathcal{C}_l^{\text{higher}} \mid \text{latency}(\mathcal{C}_l^{\text{higher}}) > \text{latency}(\mathcal{C}_l)$
\State \hspace{6mm} \textbf{\color{black}if} $\mathcal{A}_{\text{new}} \geq \mathcal{A}_m - T$ \textbf{\color{black}then} {\color{gray!80!black}\Comment{\textit{Check accuracy constraint}}}
\State \hspace{6mm} \textbf{Perform full design space exploration:}
\Statex \hspace{12mm} $\mathcal{D} = \bigcup_{l \in m} \{\mathcal{C}_l \mid \text{remaining combinations}\}$
\Statex \hspace{12mm} $\mathcal{D} \leftarrow \mathcal{D} \setminus \{\mathcal{C}_l \mid b_w, b_a \text{ are higher than current settings}\}$
\Statex \hspace{12mm} \textbf{\color{black}until} DSE reaches $I$ iterations, {\color{red!60!black}\textbf{terminate process}}\label{line:term2}
\State \hspace{6mm} \textbf{\color{black}else}
\State \hspace{12mm} Go to \textbf{step 15}
\State \hspace{6mm} \textbf{\color{black}end if}
\State \textbf{\color{black}end if}
\State \textbf{\color{black}return} $\mathcal{D}$\label{line:end}
\end{algorithmic}
\end{algorithm}

Finally, given a selected mixed-precision configuration among the obtained Pareto space, our framework evaluates the model as follows:
\begin{enumerate}[leftmargin=*]
    \item Obtains from the user the C source code, which includes the respective replacements of the original kernels with kernels incorporating the \textit{nn\_mac\_w[x]a[y]} operations.\label{item:s1}
    \item Updates the RISC-V toolchain with the new instructions, verifies their functionality in the Spike simulator, and generates the ELF executable file.\label{item:s2}
    \item Synthesizes the modified core using EDA tools for FPGA and ASIC technology, and extracts the RTL descriptions.\label{item:s3}
   % \item \blue{Simulate the post-synthesis ISA-extended Ibex processor using Verilator and along with Synopsys PrimeTime extract performance metrics and power analysis.}\label{item:s4} 
    \item Simulates the ISA-extended Ibex processor using Verilator, and, along with Synopsys PrimeTime, extracts performance metrics and power analysis.\label{item:s4} 
\end{enumerate}
Note that only step~\ref{item:s1} and~\ref{item:s4} need to be executed for different benchmarks or configurations, while steps~\ref{item:s2} and~\ref{item:s3} are executed only once.
\blue{Finally, our framework can support any DNN architecture composed of layers currently covered by our implemented kernels, including depthwise convolutional layers, dense layers, residual blocks, and pooling layers.\footnote{\blue{For more details see `\textit{How to test a new model}' section at \url{https://github.com/alexmr09/Mixed-precision-Neural-Networks-on-RISC-V-Cores}}}\label{R1C8}
}

\subsection{Voltage-scaling exploration}\label{sec:secvos}

To evaluate the impact of voltage scaling on our modified Ibex system, which integrates both the core and accelerator, we conducted a comprehensive power-performance analysis. 
The design was synthesized at a 250 MHz clock frequency, and this process ensured accurate characterization through Composite Current Source (CCS) models, under typical conditions (TT, 25°C) at 0.8V. The synthesized netlist underwent functional validation in Siemens QuestaSim 2024.3, generating VCD files later converted to SAIF for power estimation.
For voltage scaling exploration, we employed a Unified Power Format (UPF) description to define power domains, voltage levels, and retention strategies, ensuring correct functionality across multiple voltage conditions. 
Using PrimeTime, we conducted timing and power analysis at different voltage corners through \textit{define\_scaling\_lib\_group}, considering standard cell characterizations from 0.6V to 1.0V while maintaining constant temperature conditions. 
By sweeping voltage levels in 0.05V increments, we systematically analyzed power consumption and performance (GOPs/W) without extrapolation.

The results show that lowering the supply voltage significantly reduces power consumption while maintaining a constant operating frequency, leading to improved energy efficiency (GOPs/W). The total power dissipation of a circuit is given by:
\begin{equation}
P_{total} = P_{static} + a \times C \times f_s \times V_{dd}^2,
\label{eq:vdd}
\end{equation}
where $a$ is the switching activity, $C$ is the circuit's capacitance, $f_s$ is the operating clock frequency and $V_{dd}$ is the supply voltage. Since dynamic power scales quadratically with $V_{dd}$ and \blue{static} power also decreases at lower voltages, power dissipation is significantly reduced without affecting computational throughput.
However, excessive voltage scaling increases timing slack, potentially compromising system stability in more aggressive operating conditions. 
In our case, instability occurs when the supply voltage drops below 650 mV.

\begin{figure}
    \centering
    \includegraphics[width=\columnwidth]{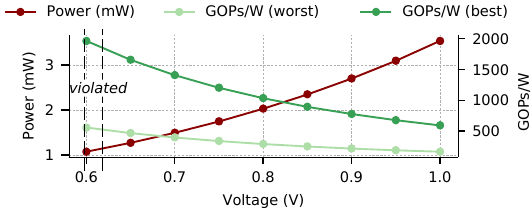}
    \caption{Impact of voltage scaling on power and energy efficiency in the modified Ibex system through EDA tools. Worst and best cases correspond to applying \textit{W8A8} and \textit{W2A2}, respectively, for the entire model.}
    \label{fig:vos}
\end{figure}

\begin{figure}
    \centering
    \includegraphics[width=0.9\linewidth]{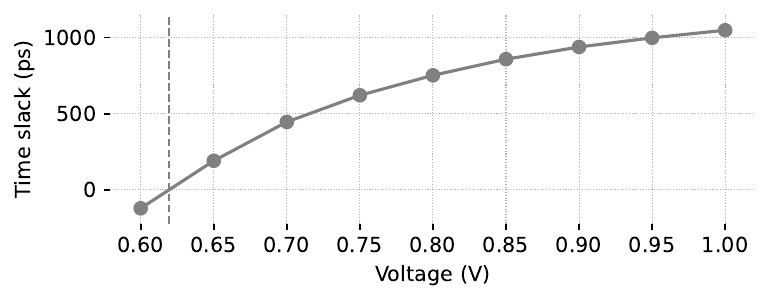}
    \caption{\blue{Time slack versus voltage for the baseline core at 250MHz, showing positive slack above 0.62V and negative slack indicating timing violations below this threshold.}}
    \label{fig:slack}
\end{figure}

In Fig.~\ref{fig:vos}, we evaluate the impact of voltage scaling on power consumption and energy efficiency. 
In this figure, we analyze the power dissipation and GOPs/W across different supply voltages ranging from 0.6V to 1.0V. 
As expected, reducing the supply voltage leads to a quadratic reduction in dynamic power. 
For instance, by lowering the voltage from 1.0V to 0.7V, power consumption decreases significantly, with a reduction of approximately 58\%. This translates directly to a $2.4\times$ increase in GOPs/W for both the worst and best cases, where the highest configuration (\textit{W8A8}) and the lowest configuration (\textit{W2A2}) are used for the entire inference, respectively. This means that the system operates more efficiently in terms of computational performance per watt. The trend continues as we further scale voltage down to 0.6V, where the lowest power levels are observed.
\blue{However, as shown in Fig.\ref{fig:slack}, excessive voltage scaling introduces timing margin violations. Specifically, time slack on the critical path reduces from +1048ps at 1.00V to -121ps at 0.60V, indicating that timing requirements at 250MHz are no longer met below 0.62V. The overall impact of complete model executions with various mixed-precision configurations is evaluated in Section~\ref{sec:soacomp}.
}\label{R2C4}

\section{Evaluation \& Results}

\subsection{Experimental Setup}

\begin{table}[t]
\setlength\tabcolsep{5pt}
\renewcommand{\arraystretch}{1.2}
\small
\caption{Description of our evaluated baseline models}
\begin{threeparttable}
\begin{tabular}{|c|c|c|c|c|}
\hline
\textbf{Model} & \textbf{Acc (\%)} & \textbf{Topology\tnote{1}} & \textbf{\#cycles} & \textbf{\#MAC} \\ \hline
CNN (CIFAR10)  & 78.9              & 3C-1D             & 110.9M            & 12.3M             \\ \hline
LeNet5         & 98.2              & 2C-3D             & 7.4M             & 423K             \\ \hline
Mcunet-vww1    & 88.9              & 1C-15R-1D         & 175.6M            & 12M             \\ \hline
MobileNetV1    & 70.4              & 14C-1D            & 5.7B              & 573M             \\ \hline
\blue{ResNet18}    & \blue{73.3}              & \blue{17C-1D}            & \blue{18.4B}              & \blue{1.8B}             \\ \hline
\end{tabular}
\begin{tablenotes}\small
\item[] $^1$Network's topology in terms of convolutional (C), dense (D), and residuals (R) layers
\vspace{-2ex}
\end{tablenotes}
\end{threeparttable}
\label{tab:baseline}
\end{table}

In this section, we evaluate the efficiency of \textit{MaRVIn} by performing an in-depth evaluation over four state-of-the-art DNNs trained on four simple and challenging image classification datasets (Table~\ref{tab:baseline}).
For our analysis, we consider: LeNet5, a CNN with 3 convolutional layers~\cite{CMSIS-NN}, MCUNet~\cite{MCUNet}, MobileNetV1~\cite{mbnet}, and ResNet18~\cite{resnet} trained on MNIST, CIFAR10, Visual Wake Words~\cite{VWW} and ImageNet dataset, respectively.
Our mixed-precision post-training quantization is based on PyTorch libraries and considers 10\% of the training dataset, while the fine-tuning process with quantization-aware training is executed using a larger portion of the same dataset and for 25 epochs (with the ability to leverage early stopping if convergence is reached sooner), on average, requiring less than 15 hours each, on an NVIDIA Tesla V100 with 32GB RAM.

For hardware results, we evaluate both ASIC and FPGA synthesis flows (Table~\ref{tab:ibex}) for the microarchitecture of the open-source Ibex core~\cite{ibexproc}, described in System Verilog.
For the FPGA workflow, we consider a Virtex-7 FPGA as a proof-of-concept platform, with the main core operating at 50MHz and the custom functional unit running at 100 MHz, and used Vivado 2023.1 for synthesis results. 
The register file is implemented using RAM32M primitives.
Cycle-accurate simulations for both performance metrics and top-1 classification accuracy are performed using Verilator~\cite{verilator}, which reads Ibex performance counters for a precise report of total cycles.
%Verilator reads the HDL code, conducts lint checks and instantiate the Verilated model. 
%Executing the resulting executable facilitates the simulation of the processor.
On the other hand, for the ASIC workflow, the processor is implemented in System Verilog RTL and synthesized with a dual-clock configuration, running at 250 MHz for the main core and 500 MHz for the functional unit. 
The design is mapped to the industrial-strength GlobalFoundries 22nm library, aligning with state-of-the-art approaches. 
Synthesis is performed using Synopsys Design Compiler with the $compile\_ultra$ command, while the register file is implemented using latches. Power analysis is conducted by leveraging the switching activity extracted from the VCD file, combined with the design's synthesized netlist, enabling accurate power estimation.

\subsection{DSE Evaluation - Latency vs Network accuracy}\label{R3C1a}\label{sec:dse_eval}

\begin{figure*}
    \centering
    \includegraphics{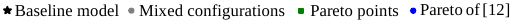}\\
    \includegraphics[width=0.22\linewidth]{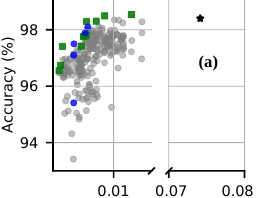}
    \includegraphics[width=0.18\linewidth]{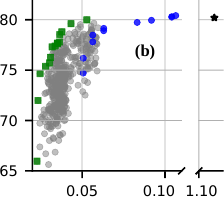}
    \includegraphics[width=0.19\linewidth]{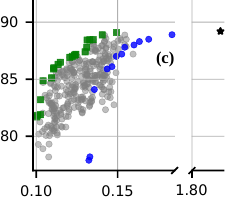}    
    \includegraphics[width=0.19\linewidth]{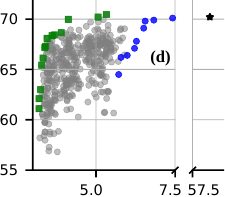}
    \includegraphics[width=0.19\linewidth]{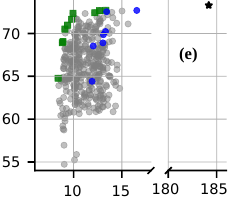}\\
    \includegraphics{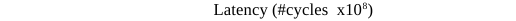}
    \caption{Pareto Space of accuracy versus latency in clock cycles, obtained from our mixed-precision configuration exploration for LeNet(a), CNN\_CIFAR10(b), MCUNet(c), MobileNetV1(d) \blue{and ResNet18(e).}}
    \label{fig:dse}\label{R1C12}
\end{figure*}

\begin{table}[t]
\setlength\tabcolsep{1.33pt}
\renewcommand{\arraystretch}{1.3}
\small
\caption{Performance comparison of our modified Ibex processor with the baseline. 
The best model per benchmark with less than 1\% accuracy loss is considered.}
\begin{threeparttable}
\resizebox{\linewidth}{!}{
\begin{tabular}{|c||cc||cc|}
\hline
\textbf{Platform} &
  \multicolumn{2}{c||}{\textbf{FPGA}} &
  \multicolumn{2}{c|}{\textbf{ASIC GF22nm}} \\ \hline
\textbf{} &
  \multicolumn{1}{c|}{\textbf{\begin{tabular}[c]{@{}c@{}}Baseline\\ Ibex\end{tabular}}} &
  \textbf{\begin{tabular}[c]{@{}c@{}}Modifed\\ Ibex\end{tabular}} &
  \multicolumn{1}{c|}{\textbf{\begin{tabular}[c]{@{}c@{}}Baseline\\ Ibex\end{tabular}}} &
  \textbf{\begin{tabular}[c]{@{}c@{}}Modifed\\ Ibex\end{tabular}} \\ \hline
\textbf{Clock Freq.} &
  \multicolumn{1}{c|}{50MHz} &
  \begin{tabular}[c]{@{}c@{}}50/\tnote{1}\\ 100MHz\end{tabular} &
  \multicolumn{1}{c|}{250MHz} &
  \begin{tabular}[c]{@{}c@{}}250/\tnote{1}\\ 500MHz\end{tabular} \\ \hline
\textbf{Precision} &
  \multicolumn{1}{c|}{32-bit} &
  2-/4-/8-b &
  \multicolumn{1}{c|}{32-bit} &
  2-/4-/8-b \\ \hline
\textbf{Power (mW)} &
  \multicolumn{1}{c|}{256\tnote{2}} &
  260\tnote{2} &
  \multicolumn{1}{c|}{1.23\tnote{3}} &
  2.03\tnote{3} /1.15\tnote{4} \\ \hline
\textbf{\begin{tabular}[c]{@{}c@{}}Resources/\\ Area\end{tabular}} &
  \multicolumn{1}{c|}{\begin{tabular}[c]{@{}c@{}}5.5K FF\\ 5.1\blue{K} LUTs\\ 4 DSPs\end{tabular}} &
  \begin{tabular}[c]{@{}c@{}}7.3K FF\\ 6.2\blue{K} LUTs\\ 4 DSPs\end{tabular} &
  \multicolumn{1}{c|}{6616$um^2$} &
  8683$um^2$ \\ \hline
\textbf{\begin{tabular}[c]{@{}c@{}}Energy\\ Efficiency\\ (GOP/s/W)\end{tabular}} &
  \multicolumn{1}{l|}{\begin{tabular}[c]{@{}l@{}}LeNet:      0.022 \\ CNN: 0.043\\ MCUNet: 0.026 \\ MbNetV1: 0.04   \\ \blue{ResNet18: 0.038}\end{tabular}} &
  \begin{tabular}[c]{@{}c@{}}0.438\\ 1.09\\ 0.34\\ 0.565 \\ \blue{0.538}\end{tabular} &
  \multicolumn{1}{c|}{\begin{tabular}[c]{@{}c@{}} 23.1\\ 45.2\\ 26.8 \\ 40.9 \\ \blue{40}\end{tabular}} &
  \begin{tabular}[c]{@{}c@{}} 280\tnote{3}  / \blue{496\tnote{4}}  \\ 831\tnote{3}  / \blue{1471\tnote{4}}\hspace{0.2em}  \\ 237\tnote{3}  / \blue{385\tnote{4}}  \\ 362\tnote{3}  / \blue{639\tnote{4}}  \\ \blue{344\tnote{3}  / \blue{609\tnote{4}} }\end{tabular} \\ \hline
\end{tabular}
}
\begin{tablenotes}\small
\item[] $^1$Basic core's clock frequency / Multi-pumped units' clock frequency \xspace \xspace $^2$ 28\% \blue{static}\label{R2C8} power \xspace \xspace$^3$ 0.8V operating voltage \item[] \blue{$^4$ 0.62V operating voltage}
\vspace{-2ex}
\end{tablenotes}
\end{threeparttable}
\label{tab:ibex}\vspace{-2ex}\label{R1C11}\label{R1C10}
\end{table}

As mentioned, our framework starts with a design space exploration to find different mixed-precision configurations and pruning variations for each input model.
Fig.~\ref{fig:dse} presents the Pareto space between accuracy and required number of clock cycles for all the DNNs examined.
\blue{These clock cycles correspond to the end-to-end execution of the entire network.}\label{R1C6}
In Fig.~\ref{fig:dse}, the black star represents the pre-trained baseline model.
The gray circles are the quantized models with mixed precision among their conventional or pruned layers, while the green squares form the Pareto front.
The blue circles are the Pareto front of our previous version~\cite{iccad24mixed}, i.e., without adapting the bit-width of activation and without incorporating our pruning approach.
\blue{This evaluation also includes the minor latency overheads (ranging from 0.003\% to 1.6\% of the total execution cycles) introduced by the pre-processing and operand packing performed at the C level. These steps are fully automated, executed offline, and required only once per model.}\label{R1C5a}
Note that, compared to our prior version, we integrate Verilator into the DSE to accurately measure cycle counts instead of relying on the number of MAC operations.
This adjustment is necessary because determining latency-optimal configurations per layer becomes now a non-trivial problem for our current design space, due to the expanded mixed-precision options (e.g., \textit{W8A2} vs \textit{W4A4}) and varying pruning rates.
Furthermore, for the DSE of all models, we set a 1\% accuracy constraint, a maximum pruning rate of $p_{\text{max}} = 50\%$, and a total of $I = 500$ iterations.
As derived, the number of models explored depends on the number of layers.
In total, to generate Fig.~\ref{fig:dse}, we evaluated more than \blue{2000} quantized models.
It is observed that different precisions in layers ranging between 2,4 and 8 bits can achieve accuracies comparable to or the same as the full-precision counterparts.

For all benchmarks, more than \blue{92\%} reduction in the number of cycle counts can be achieved in their mixed quantized derivatives for less than 1\% accuracy loss, while this number increases to 95\% when up to 5\% accuracy loss is applicable.
Moreover, compared to~\cite{iccad24mixed}, our enhanced exploration can achieve much better performance, since for less than 1\% accuracy loss, the inference time is almost 32\% better, on average for all models.

\subsection{Per-technique analysis}\label{sec:pertech}

\begin{figure}
    \centering
    \includegraphics[width=\columnwidth]{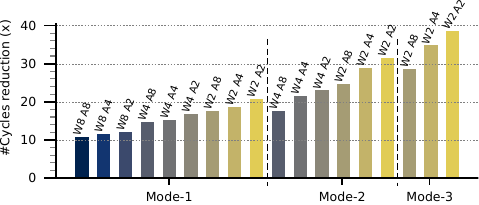}
    \caption{Cycle count reduction for a convolution layer when using each \textit{Mode}, compared to the baseline. The impact of individual mixed-precision configurations, corresponding to a single optimization per \textit{Mode}, is illustrated.}
    \label{fig:modeeval}
\end{figure}

\begin{figure*}
    \centering
    \includegraphics[width=\textwidth]{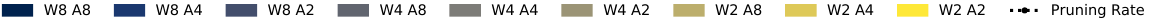}\\
    \includegraphics{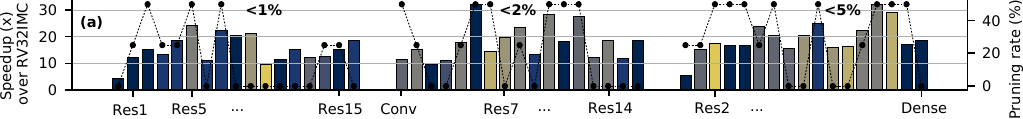}\\\vspace{1ex}
    \includegraphics{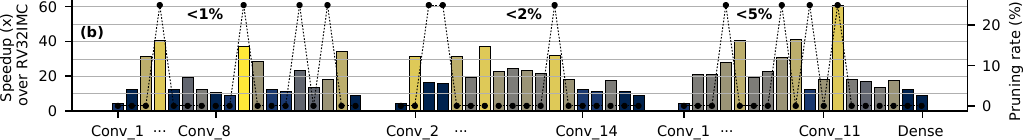}     
    \includegraphics{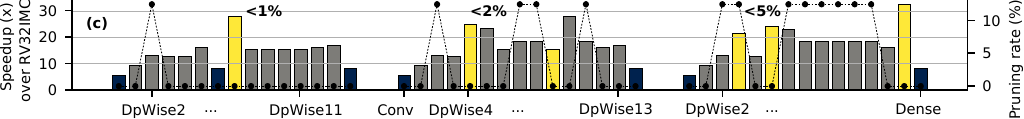}\\
    \includegraphics{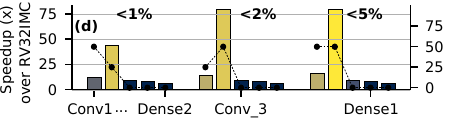} 
    \includegraphics{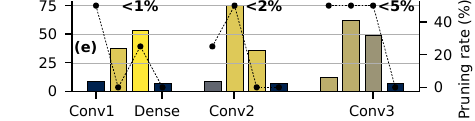}
    \caption{Speedup gains compared to the baseline Ibex RV32IMC for 4 different benchmarks, i.e., MCUNet (a), \blue{ResNet18 (b)}, MobileNetV1 \blue{(c)}, CIFAR10 CNN \blue{(d)}, LeNet \blue{(e)}, are shown. For each benchmark, three different mixed-precision models have been selected, \blue{each one satisfying a different accuracy loss (1\%, 2\%, 5\%)}. The selected pruning rate per layer, expressed as a percentage, is depicted on the right Y-axis.}
    \label{fig:speedups}\label{R3C1b}
\end{figure*}

The impact of each distinct optimization technique (different \textit{Mode}) that is applied within a processor execution is evaluated in Fig.~\ref{fig:modeeval}.
To generate Fig.~\ref{fig:modeeval} we consider as an example one distinct convolutional layer, i.e., the second layer of the CIFAR10 CNN, and evaluate the relative speedups of the standalone \textit{Mode-1,-2,-3}, in terms of number of cycles reduction.
In other words, we report the gains of each technique as if it was solely applied throughout the whole layer and examined for all the nine different bit-widths.
As can be noticed, not every single \textit{mode} supports all nine mixed-precision configurations. 
It is observed that \textit{Mode-1}, due to the high parallelization of 8-bit operands and minimization of loads and stores, can achieve on average $14\times$ speedup compared to the baseline RV32IMC, and almost $21\times$ when 2-bit weights and activations are utilized.
On the other hand, when the standalone multi-pumping technique is applied (\textit{Mode-2}), it can increase the average speedup to $24\times$.
Finally, when 2-bit weights are considered (\textit{Mode-3}), our soft SIMD approach obtains an extra speedup, having a significant reduction in total cycles of almost $34\times$.
Note that this number is increased by $3\times$ compared to our previous work~\cite{iccad24mixed}, where the bit-width of activations was kept fixed.

\blue{Despite the significant speedups, each mode introduces some area and power overhead on top of its previous mode.
Specifically, to support mode-1, the modified core introduces a 22\% power overhead over the baseline (1.23mW), primarily due to added parallelism. 
Mode-2 incurs the highest power increase (25\% on top of mode-1) because of the multi-pumping technique that doubles the ALU frequency, while mode-3 and soft SIMD add an additional 7\%.
In terms of area, mode-1 introduces the largest overhead (19\%) due to extra logic, while mode-2 and mode-3 add just 4\% and 5\%, respectively, as they rely mainly on algorithmic techniques.
Regarding the FPGA design, mode-1 increases area by 26\% in LUTs and 11\% in FFs. Mode-2 adds 2\% LUTs and 6\% FFs over mode-1, while mode-3 contributes an additional 9\% LUTs and 2\% FFs. DSP usage remains constant across all modes.
The total cumulative overheads can be shown in Table~\ref{tab:ibex}.
}\label{R1C4b}

As mentioned earlier, the sensitivity and error resilience of each layer can vary, and hence the top-1 models' accuracy is tightly dependent on the selected precision for each layer. 
Subsequently, in order to put the delivered trade-offs between total execution time of DNN inference and top-1 accuracy into perspective, and analyze the contribution of each layer and each configuration, we generate Fig.~\ref{fig:speedups}.
This figure evaluates three user-defined accuracy loss thresholds and extracts corresponding latency-optimal mixed-precision configurations from the DSE, each minimizing latency for its respective accuracy constraint.
The bit-width configurations of weights and activations per layer are depicted using distinct colors, while the right Y-axis shows the pruning ratio for each layer.
Then, the speedup for all layers derived from Verilator is reported.
The key observations regarding the mixed precision selection are two: 1) the less complex models like LeNet and CIFAR10 CNN allow aggressive quantization down to 2 bits (especially in weights) in most of their layers with minimal accuracy degradation ($<1\%$) and thus able to achieve the maximum possible performance gains, and 2) although the more challenging \blue{MobileNetV1, ResNet18} and MCUNet models barely utilize the beneficial \textit{W2A2} configuration, they manage to map most of their layers to \textit{W4A4} option with small impact on their accuracy ($<2\%$).
Overall and on average for all layers, our proposed ISA extensions can achieve from \blue{17.6}$\times$ up to \blue{23.3}$\times$ speedup for 1\% up to 5\% accuracy degradation, respectively.
It is also noteworthy that MCUNet presents less significant gains compared to other benchmarks due to the high amount of depthwise convolutions.
The latter do not enable the same degree of data reuse as in standard \& point-wise convolutions, while also they differ in the overheads (e.g., branch instructions) they introduce.
\blue{Finally, these consistent performance gains indicate that our proposed framework could scale effectively from simple models to more complex ones, as it leverages 2-, 4-, and 8-bit precision, which has been shown to preserve high inference accuracy even in larger architectures~\cite{csurarme}.}\label{R3C1c}

\subsection{Comparison against state-of-the-art}\label{sec:soacomp}

% \usepackage{multirow}
\begin{table*}[ht]
\setlength\tabcolsep{0.8pt}
\renewcommand{\arraystretch}{1.3}
\caption{Comparison with state-of-the-art solutions. Performance metrics \blue{along with precision configurations are presented. Efficiency reflects the raw performance of a typical convolutional layer, using only our least efficient to most efficient mode.}}
\begin{threeparttable}
\begin{tabular}{cc|ccccccc|c|c}
\hline
\textbf{} &
  \textbf{} &
  \multicolumn{1}{c|}{\textbf{TC'24\cite{wang2024optimizing}}} &
  \multicolumn{1}{c|}{\textbf{HPCA'23\cite{Mix-GEMM}}} &
  \multicolumn{1}{c|}{\textbf{ISVLSI'20\cite{ottavi2020mixed}}} &
  \multicolumn{1}{c|}{\textbf{JSSC'18\cite{lee2018unpu}}} &
  \multicolumn{1}{c|}{\textbf{TCAD'20\cite{tcad20}}} &
  \multicolumn{1}{c|}{\textbf{DATE'20\cite{XpulpNN}}} &
  \multicolumn{1}{c|}{\textbf{TVLSI'24\cite{10705106}}} &
  \multicolumn{1}{l|}{\textbf{ICCAD'24\cite{iccad24mixed}}} &
  \textbf{\textit{MaRVIn}} \\ \hline
\multicolumn{1}{c|}{\multirow{4}{*}{\textbf{\rotatebox{90}{\begin{tabular}[c]{@{}c@{}}Archite-\\ cture\end{tabular}}}}} &
  \textbf{Platform} &
  90nm &
  22nm &
  22nm &
  65nm &
  65nm &
  22nm &
  28nm &
  7nm &
  22nm \\ \cline{2-11} 
\multicolumn{1}{c|}{} &
  \textbf{Precision} &
  32 b W/A&
  2-8 b W/A&
  2/4/8 b W/A&
  1-16 b W&
  16 b W/A&
  2/4/8 b W/A&
  4/8/16 W/A &
  2/4/8 b W &
  2/4/8 b W/A \\ \cline{2-11} 
\multicolumn{1}{c|}{} &
  \textbf{Clk Freq.} &
  100MHz &
  1.2GHz &
  250MHz &
  200MHz &
  200MHz &
  600MHz &
  1.05GHz &
  250MHz &
  250MHz \\ \cline{2-11} 
\multicolumn{1}{c|}{} &
  \textbf{Area/Power} &
  \begin{tabular}[c]{@{}c@{}}6.44$mm^2$/ \\ 5.8mW\end{tabular} &
  \begin{tabular}[c]{@{}c@{}}0.014$mm^2$/\\ 9.9mW\tnote{3}\end{tabular} &
  \begin{tabular}[c]{@{}c@{}}0.002$mm^2$/\\ 5.5mW\tnote{3}\end{tabular} &
  \begin{tabular}[c]{@{}c@{}}16$mm^2$/\\ 288mW\end{tabular} &
  \begin{tabular}[c]{@{}c@{}}11.47$mm^2$/\\ 805mW\end{tabular} &
  \begin{tabular}[c]{@{}c@{}}0.04$mm^2$/\\ 43.5mW\tnote{3}\end{tabular} &
  \begin{tabular}[c]{@{}c@{}}1.20$mm^2$/\\ 533mW\end{tabular} &
  \begin{tabular}[c]{@{}c@{}}0.038$mm^2$/\\ 0.58mW\end{tabular} &
  \begin{tabular}[c]{@{}c@{}}0.009$mm^2$/\\ 1.15mW\end{tabular} \\ \hline
\multicolumn{1}{c|}{\multirow{2}{*}{\rotatebox{90}{\textbf{\hspace{-9ex}\begin{tabular}[c]{@{}c@{}}Performance\\ \end{tabular}}}}} &  
  \textbf{GOPs} &
  0.23 &
  11.9 &
  3.3 &
  514 &
  288 &
  47.9 &
  343.1-737.9 &
  0.24-0.85 &
  0.6-2.1 \\ \cline{2-11} 
\multicolumn{1}{c|}{} &
  \textbf{\begin{tabular}[c]{@{}c@{}}Area Eff.\\ (GOPs/$mm^2$)\end{tabular}} &
  0.04 &
  850 &
  1650 &
  32.1 &
  25.1 &
  1198 &
  285.8-614.6 &
  6.3-22.4 &
  66.7-233.3 \\ \cline{2-11} 
\multicolumn{1}{c|}{} &
  \textbf{\begin{tabular}[c]{@{}c@{}}Energy Eff.\\ (GOPs/W)\end{tabular}} &
  38.8\tnote{1} &
  500-1166\tnote{2} &
  200-600\tnote{1} &
  1750\tnote{2} &
  357.8\tnote{2} &
  700-1100\tnote{1} &
  643-1383\tnote{1} &
  415-1470\tnote{2} &
  522-1836\tnote{1,4} \\ \hline
\end{tabular}
\begin{tablenotes}\small
\item[] $^1$Throughput and energy efficiency were calculated based on a typical convolution layer $^3$Area includes only extended units
\item[] $^2$Throughput and energy efficiency were calculated based on the average values of examined DNNs
\item[] $^4$\blue{GOPs/W for W8I8: 522, W8I4: 546, W8I2: 572, W4I8: 837, W4I4: 1029, W4I2: 1094, W2I8: 1369, W2I4: 1658, W2I2: 1836}
\end{tablenotes}
\end{threeparttable}
\label{tab:soa}
\end{table*}

In this section, we evaluate \textit{MaRVIn} against the most relevant state-of-the-art solutions and our previous version, highlighting our incremental improvements achieved. 
We note that our solution is synthesized and evaluated using GlobalFoundries 22nm FDX, an industrial-strength technology, ensuring a more realistic assessment compared to academic PDKs such as the ASAP 7nm library used in our previous work and others.
We specifically examine hardware-software co-designed architectures for executing DNNs on CPU-based platforms, incorporating ISA extensions and custom functional units. A detailed comparative analysis is provided in Table~\ref{tab:soa}.
To assess the performance of our modified RISC-V processor with custom ISA extensions against state-of-the-art solutions, we report the peak performance values stated in each study. For a fair comparison, we consider either a representative convolution layer or the average peak performance across all evaluated DNNs.
Some prior works do not account for accuracy degradation in their reported peak performance—sometimes leading to significant trade-offs~\cite{Mix-GEMM}—and hence, this aspect is not analyzed here. In contrast, our evaluation presents an energy efficiency range with an accuracy deviation of less than 1\%, ensuring near floating-point performance.

As presented in Table~\ref{tab:soa}, our energy efficiency values correspond to the lowest and highest efficiency achieved \blue{across a typical convolution layer, while Table~\ref{tab:ibex} presents the efficiency of each examined benchmark.}
The highest efficiency is observed at 0.62V, the lowest operating voltage where the core design remains within clock constraints.
As shown, our lowest energy efficiency, on average (\blue{720} GOPs/W), exceeds that of MIX-GEMM~\cite{Mix-GEMM} when maintaining accuracy degradation below 1\%. 
Compared to \cite{ottavi2020mixed}, our efficiency ranges from 0.9$\times$ up to 9.2$\times$ higher, depending on the configuration.
Focusing on MobileNetV1, MIX-GEMM achieves approximately 500 GOPs/W with minor accuracy degradation, whereas our approach surpasses this, reaching 639 GOPs/W and scaling up to 801 GOPs/W with an accuracy loss of up to 5\%.
Similarly, XPulpNN~\cite{XpulpNN}, which integrates SIMD units capable of executing 4 MACs (8-bit) to 16 MACs (2-bit) per cycle, achieves competitive performance on a single convolution layer. 
However, when evaluated on our full CNN model \blue{(1471 GOPs/W)}, its energy efficiency remains 25\% lower than that of our processor.
Additionally, UNPU~\cite{lee2018unpu}, despite its high-performance decoupled accelerator, faces limitations due to reduced flexibility. The necessity for complex software stacks in such accelerators often demands specialized offloading mechanisms at the hardware or software level, which can impact overall throughput.
% It is important to note that Table~\ref{tab:soa} reports only designs constrained to 1\% accuracy loss.
Note also that when prioritizing energy efficiency over top-1 accuracy, our processor reaches an average of \blue{905} GOPs/W, with a peak performance of nearly 1.8 TOPs/W on the CIFAR10 CNN model under a 5\% accuracy degradation.
\blue{
Specifically, the energy efficiency of the best designs satisfying the 2\% accuracy loss criteria is 796 GOPs/W averaged across all benchmarks.\label{R1C9}
}
SPEED~\cite{10705106} achieves a high operating frequency of 1.05 GHz through precise pipeline stage allocation, but this comes at the cost of reduced area and energy efficiency compared to \textit{MaRVIn}.
Finally, while our current implementation achieves slightly lower energy efficiency than our previous work~\cite{iccad24mixed}, due to differences in technology libraries and variations in the power characteristics of the standard cells, it significantly improves raw computational throughput-the prime focus of our work.
On average, \textit{MaRVIn} delivers a $2.5\times$ increase in GOPs compared to its previous version.
\section{conclusion}
In this work, we embrace the power of mixed-precision neural networks to address the limitations of conventional CPU architectures in managing precision efficiently.
We propose enhancements in a generic RISC-V CPU core with lightweight hardware modifications and an end-to-end automated framework, \textit{MaRVIn}, for generating and evaluating mixed-precision models.
We extend the RISC-V ISA with nine novel instructions, enabling multiple operational modes that optimize computations through operand packing, multi-pumping, and a soft SIMD technique.
By integrating these enhancements into a RISC-V-based processor, we demonstrate that accuracy can be maintained while achieving significant improvements in energy efficiency and performance, marking a major step toward energy-efficient computing for resource-constrained devices.

\bibliographystyle{IEEEtran}
\bibliography{references}

\vfill
\begin{IEEEbiography}[
{\includegraphics[width=1in,height=1.35in,clip,keepaspectratio]{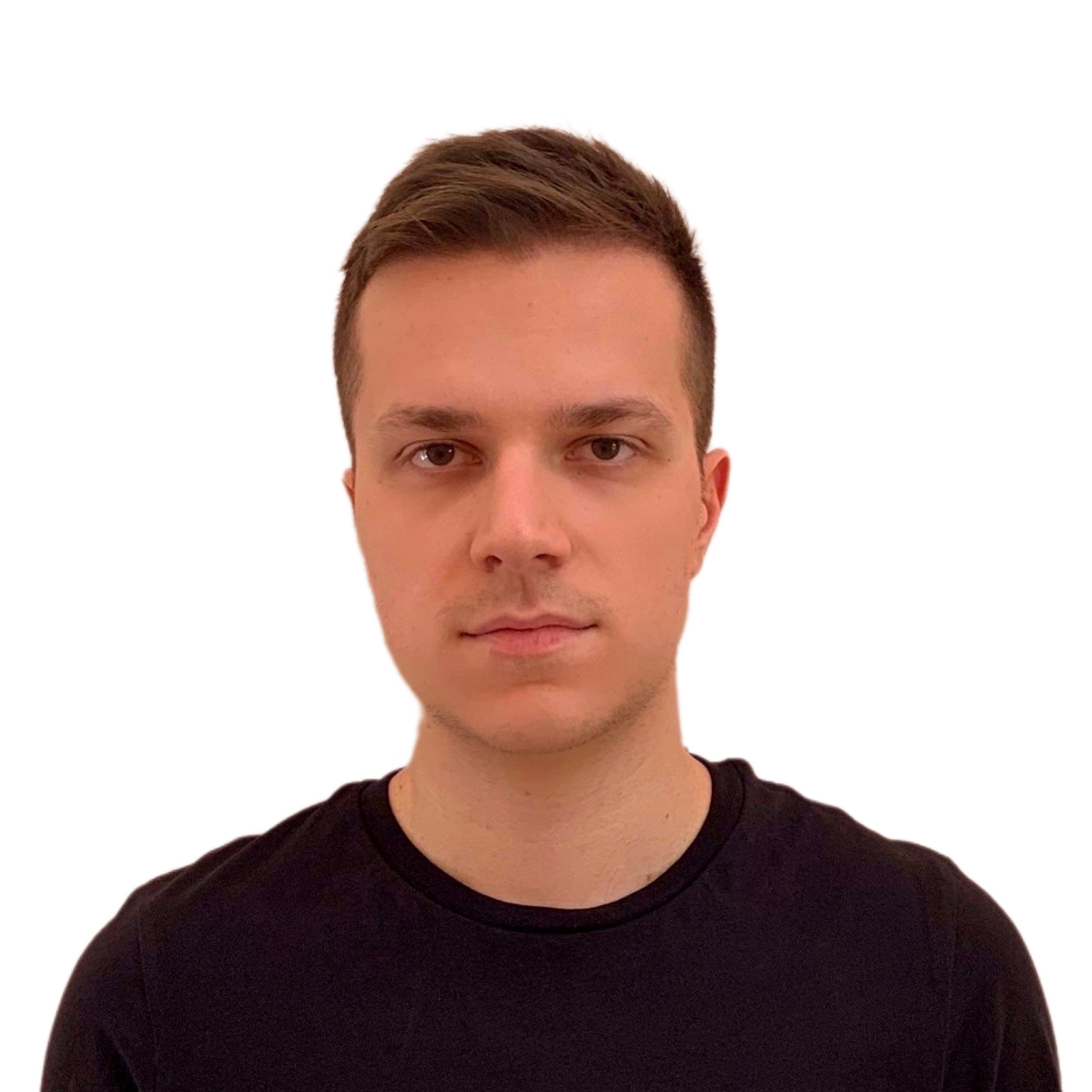}}
] {Giorgos Armeniakos}
(Member, IEEE) received the Diploma degree from the Department of Electrical and Computer Engineering (ECE), National Technical University of Athens (NTUA), Greece, in 2020, where he is currently pursuing the PhD degree. His research interests include approximate computing, digital circuit design, low power design, machine learning, telecommunications, and optimizations. He has worked as a computer engineer on several research ESA and Horizon projects and has published more than 15 papers. He holds one best paper award and one best paper nomination for his works on digital approximate designs.
\end{IEEEbiography}

\begin{IEEEbiography}[{\includegraphics[width=1in,height=1.35in,clip,keepaspectratio]{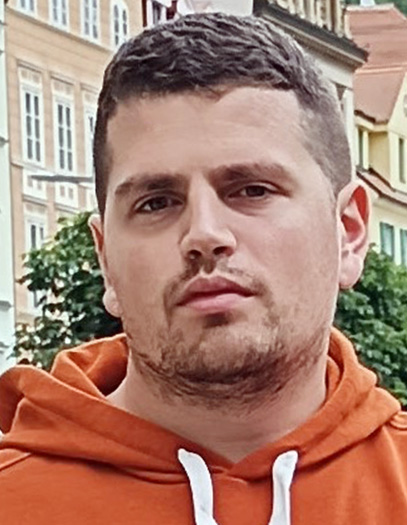}}] {Alexis Maras}
 received his Diploma in Electrical and Computer Engineering from the National Technical University of Athens (NTUA) in 2024. He is currently pursuing a PhD at the Microprocessors and Digital Systems Lab (NTUA). His research focuses on software-hardware co-optimization for machine learning applications, with an emphasis on design space exploration for dataflow optimization and the development of efficient hardware accelerators.
\end{IEEEbiography}

\begin{IEEEbiography}[{\includegraphics[width=1in,height=1.35in,clip,keepaspectratio]{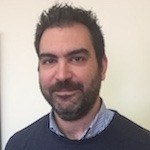}}]{Sotirios Xydis} (Member, IEEE) received the PhD degree in electrical and computer engineering from
the National Technical University of Athens (NTUA),
Greece, in 2011. He is currently an assistant professor
with the National Technical University of Athens,
Department of Electrical and Computer Engineering.
His research interests include energy efficient computing, design space exploration, high-level synthesis, accelerator design. He has published more than
100 papers in international journals and conferences and
three best paper awards.
\end{IEEEbiography}

\begin{IEEEbiography}[{\includegraphics[width=1in,height=1.25in,clip,keepaspectratio]{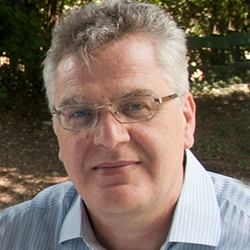}}] {Dimitrios Soudris} (Member, IEEE) received
the Diploma and Ph.D. degrees in electrical engineering from the University of Patras, Patras,
Greece, in 1987 and 1992, respectively. Since
1995, he has been a Professor with the Department of Electrical and Computer Engineering,
Democritus University of Thrace, Xanthi, Greece.
He is currently a Professor with the School of
Electrical and Computer Engineering, National
Technical University of Athens, Athens, Greece.
He has authored or coauthored more than 600 papers in international journals/conferences. He has coauthored/coedited seven Kluwer/Springer books.
He is also the leader and a principal investigator in research projects funded
by Greek Government and Industry, European Commission, ENIAC-JU,
and European Space Agency. His current research interests include high performance computing, embedded systems, reconfigurable architectures,
reliability, and low-power VLSI design. He was a recipient of the award
from INTEL and IBM for EU Project LPGD 25256; ASP-DAC 05 and VLSI
05 awards for EU AMDREL IST-2001-34379, and several HiPEAC awards.
He has served as the general/program chair in several conferences.
\end{IEEEbiography}

\end{document}